%% file: main.tex
\documentclass{article}

\usepackage[preprint]{neurips_2024}

\usepackage[utf8]{inputenc} %
\usepackage[T1]{fontenc}    %
\usepackage{hyperref}       %
\usepackage{url}            %
\usepackage{booktabs}       %
\usepackage{amsfonts}       %
\usepackage{nicefrac}       %
\usepackage{microtype}      %
\usepackage{xcolor}         %

\usepackage{tikz} %
\usepackage{multirow}
\usepackage{url}
\usepackage{graphicx}
\usepackage{wrapfig}
\usepackage{subcaption}
\usepackage{tcolorbox}
\usepackage{mathtools}
\usepackage{algorithm}
\usepackage{algpseudocode}
\usepackage{stackengine}
\usepackage{selectp}

\DeclarePairedDelimiter\floor{\lfloor}{\rfloor}

\title{Attention as an RNN}

\author{%
  Leo Feng
\\
  Mila \& Borealis AI\\
  \texttt{leo.feng@mila.quebec} \\
   \And
  Frederick Tung \\
  Borealis AI \\
  \texttt{frederick.tung@borealisai.com}
  \And
  Hossein Hajimirsadeghi \\
   Borealis AI \\
  \texttt{hossein.hajimirsadeghi@borealisai.com} \\
   \And
   Mohamed Osama Ahmed\\
   Borealis AI \\
  \texttt{mohamed.o.ahmed@borealisai.com} \\
   \And
   Yoshua Bengio \\
   Mila -- Université de Montréal \\
  \texttt{yoshua.bengio@mila.quebec} \\
   \And
   Greg Mori\\
   Borealis AI \\
  \texttt{greg.mori@borealisai.com} \\
}

\begin{document}

\maketitle

\begin{abstract}
The advent of Transformers marked a significant breakthrough in sequence modelling, providing a highly performant architecture capable of leveraging GPU parallelism. 
However, Transformers are computationally expensive at inference time, limiting their applications, particularly in low-resource settings (e.g., mobile and embedded devices). 
Addressing this, we (1) begin by showing that attention can be viewed as a special Recurrent Neural Network (RNN) with the ability to compute its \textit{many-to-one} RNN output efficiently. 
We then (2) show that popular attention-based models such as Transformers can be viewed as RNN variants. 
However, unlike traditional RNNs (e.g., LSTMs), these models cannot be updated efficiently with new tokens, an important property in sequence modelling. 
Tackling this, we (3) introduce a new efficient method of computing attention's \textit{many-to-many} RNN output based on the parallel prefix scan algorithm. 
Building on the new attention formulation, we (4) introduce \textbf{Aaren}, an attention-based module that can not only (i) be trained in parallel (like Transformers) but also (ii) be updated efficiently with new tokens, requiring only constant memory for inferences (like traditional RNNs).
Empirically, we show Aarens achieve comparable performance to Transformers on $38$ datasets spread across four popular sequential problem settings: reinforcement learning, event forecasting, time series classification, and time series forecasting tasks while being more time and memory-efficient.

\end{abstract}

\input{core/00_introduction.tex}

\input{core/01_background.tex}

\input{core/02_methodology.tex}

\input{core/03_experiments.tex}

\input{core/04_relatedwork.tex}

\input{core/05_conclusion.tex}

\newpage

\bibliographystyle{apalike}
\bibliography{neurips_2024}

\newpage
\appendix

\input{core/99_appendix.tex}

\end{document}

%% file: core/00_introduction.tex
\section{Introduction} \label{sec:introduction}

Advancements in sequence modelling are highly impactful due to the wide range of applications, including reinforcement learning (e.g., robotics and autonomous driving), time series classification (e.g., financial fraud detection and medical diagnoses), and time series forecasting (e.g., weather and energy consumption predictions).
Over the past several years, an extensively explored topic in sequence modelling is that of Transformer-based models~\citep{vaswani2017attention}. 
This is due to Transformers' strong performance and ability to leverage GPU parallelism. 
As a result, numerous Transformer-specific survey papers have been written for various sequential settings such as reinforcement learning~\citep{agarwal2023transformers,li2023survey}, time series~\citep{lin2022survey, jiang2024empowering}, and speech processing~\citep{latif2023transformers}. 

With the rapid increase in low-resource domains such as battery-powered devices, deployed models in these domains must be computationally efficient. 
Transformers, on the other hand, are expensive due to their quadratic scaling in memory and computation.
Although their efficiency can be enhanced at inference time using techniques such as KV-caching~\citep{pope2023efficiently}, Transformers remain expensive for low-resource domains due to requiring (1) linear memory in the number of tokens and (2) the caching of all preceding tokens to the model. 
The effect of these limitations is exacerbated in settings with long contexts (i.e., large number of tokens) such as those common in time series (e.g., climate and economics).

To address this issue, we begin by examining \textit{attention}, the component contributing to Transformers quadratic computational complexity. We show that (1) attention can be viewed as a special Recurrent Neural Network (RNN) with the ability to compute its \textit{many-to-one} RNN output efficiently. 
Leveraging the RNN formulation of attention, we (2) show that popular attention-based models (e.g., Transformers and Perceivers) can be viewed as RNNs.  
However, unlike traditional RNNs such as LSTMs~\citep{hochreiter1997lstm} and GRUs~\citep{cho2014gru}, these attention-based models are unable to perform efficient updates with new tokens, an important property in sequential problem settings where data is processed in a stream.

To address this, we (3) introduce a new formulation of attention based on the parallel prefix scan algorithm~\citep{blelloch1990prefix} that efficiently computes attention's \textit{many-to-many} RNN output. 
Building on this new attention formulation, we (4) introduce \textbf{Aaren} (\textbf{[A]}ttention \textbf{[a]}s a \textbf{[re]}current neural \textbf{[n]}etwork), a computationally efficient module that can not only (i) be trained in parallel (like Transformers) but also (ii) be efficiently updated with new tokens, requiring only constant memory for inferences (like traditional RNNs).
Empirically, we show Aarens achieve comparable performance to Transformers on $38$ datasets spread across four popular sequential data settings: reinforcement learning, event forecasting, time series classification, and time series forecasting tasks while being more time and memory-efficient.

%% file: core/01_background.tex
\section{Background} \label{sec:background}

\subsection{Recurrent Neural Networks}

Recurrent Neural Networks (RNNs) are specialized models for sequence modelling. 
In brief, RNNs process sequential data by iteratively computing hidden states as follows:
$$h_{t} = f_\theta(h_{t-1}, x_t)$$
where $t$ denotes the step index, $x$ represents a token, $h$ represents the hidden state, and $f_\theta$ is a neural network parameterized by $\theta$. 
The value of the initial hidden state $h_0$ is typically learned via backpropagation. 
Popular RNNs such as LSTMs~\citep{hochreiter1997lstm} and GRUs~\citep{cho2014gru} fix the size of the hidden states $h_t$ to a constant irrespective of the step index.
As such, these models are efficient at test time, requiring only constant memory and time per token and can be easily updated with new tokens, an important property in sequence modelling.
However, due to being iterative by design, these popular RNNs also suffer scalability issues due to their lack of parallelizability. 
As such, RNNs were replaced by a parallelizable attention-based module, Transformers~\citep{vaswani2017attention}, as the standard for many sequence modelling settings.

\subsection{Attention}

Attention retrieves information from a set of $X_C$ context tokens for a given set of query tokens $X_Q$ as follows:
$$\mathrm{Attention}(Q, K, V) = \mathrm{softmax}(Q K^T)V
$$
where $Q = X_Q W_q$ is the query matrix, $K = X_C W_k$ is the key matrix, and $V = X_C W_v$ is the value matrix. 
$W_q, W_k, W_v \in \mathbb{R}^{d \times d}$ are weight matrices (learned parameters). 
$\mathrm{softmax}(QK^T)$ computes weights of the context tokens for a weighted average. 
Notably, unlike RNNs, attention is not designed to be iterative; instead, it is designed to easily leverage GPU parallelism. 
Transformers~\citep{vaswani2017attention} use self-attention\footnote{For sequential data, a causal mask is typically used to prevent attending to future timesteps.}, a special case of attention, where the query tokens are the same as the context tokens. 
However, self-attention requires quadratic computation with respect to the number of tokens and is not efficiently updateable with new tokens. 
As a result, Transformers are computationally expensive, limiting their applications in low-resource domains.

%% file: core/02_methodology.tex
\section{Methodology} \label{sec:method}

\begin{figure}[ht]
    \centering
    
    \begin{subfigure}[t]{0.32\textwidth}
        \centering
        \includegraphics[height=1.0in]{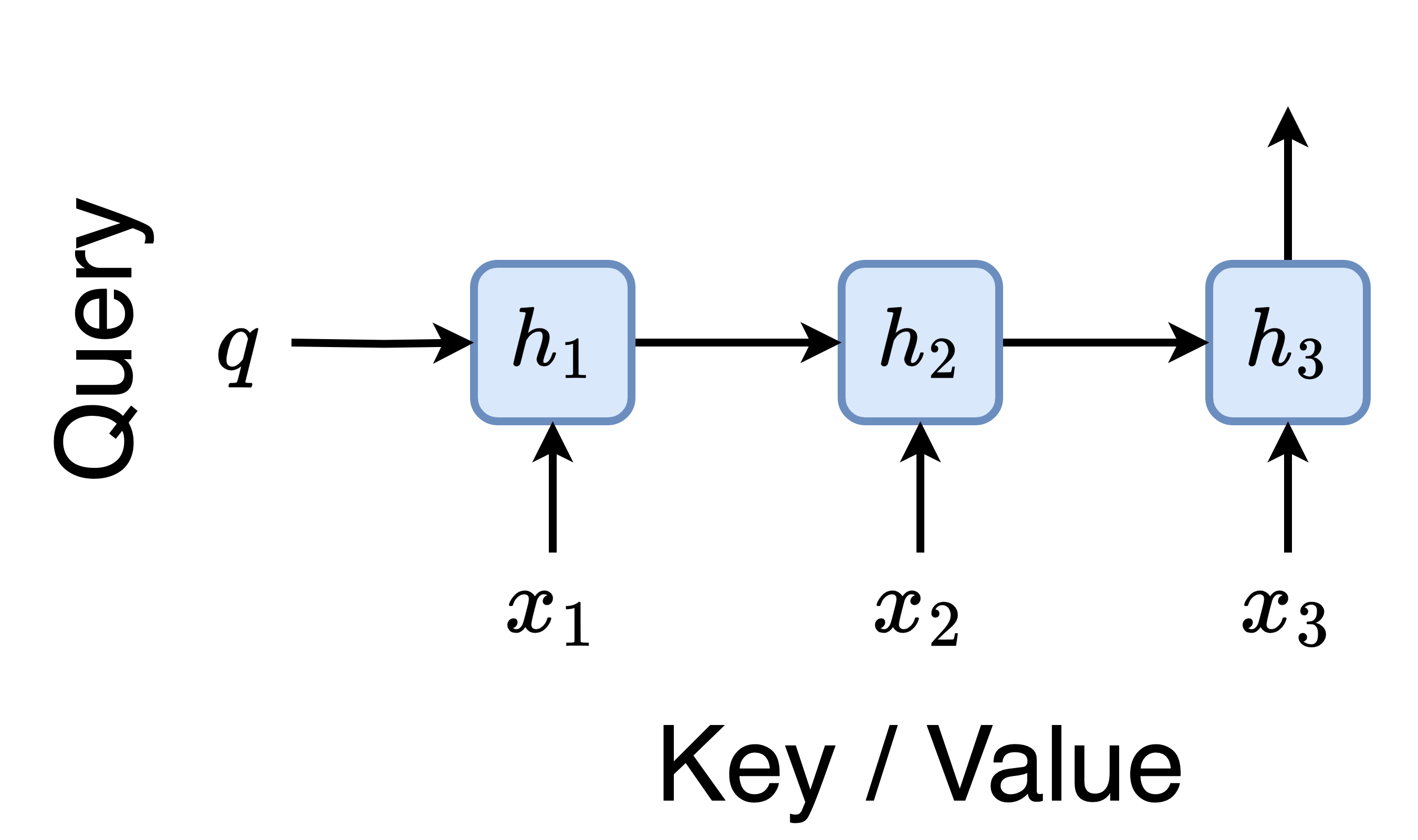}
        \caption{Conventional Attention}
        \label{fig:many_to_one_AttRNN}
    \end{subfigure}
    \begin{subfigure}[t]{0.32\textwidth}
        \centering
        \includegraphics[height=1.0in]{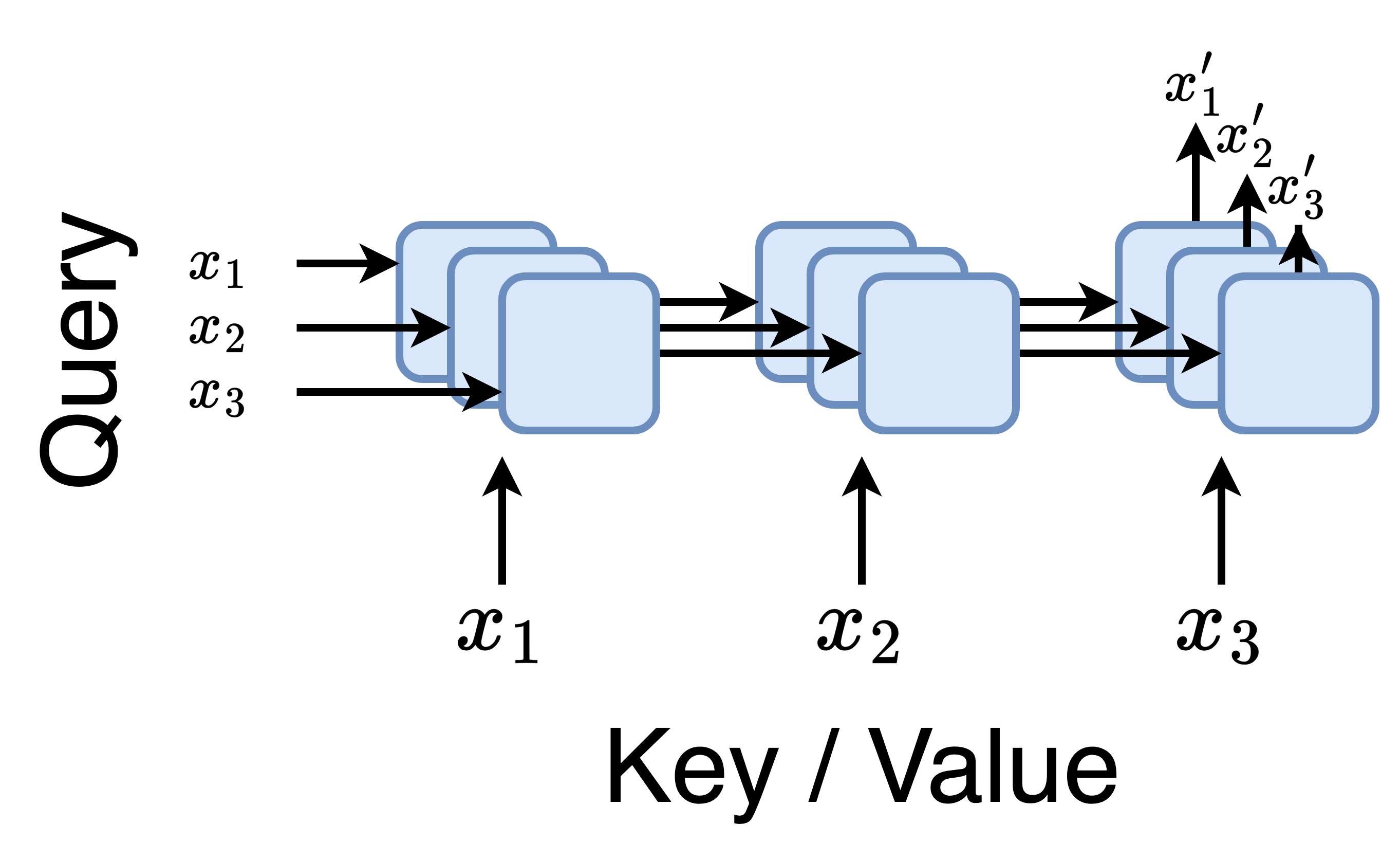}
        \caption{Transformer's Self-Attention}
        \label{fig:causal-transformer}
    \end{subfigure}
    \begin{subfigure}[t]{0.32\textwidth}
        \centering
        \includegraphics[height=1.0in]{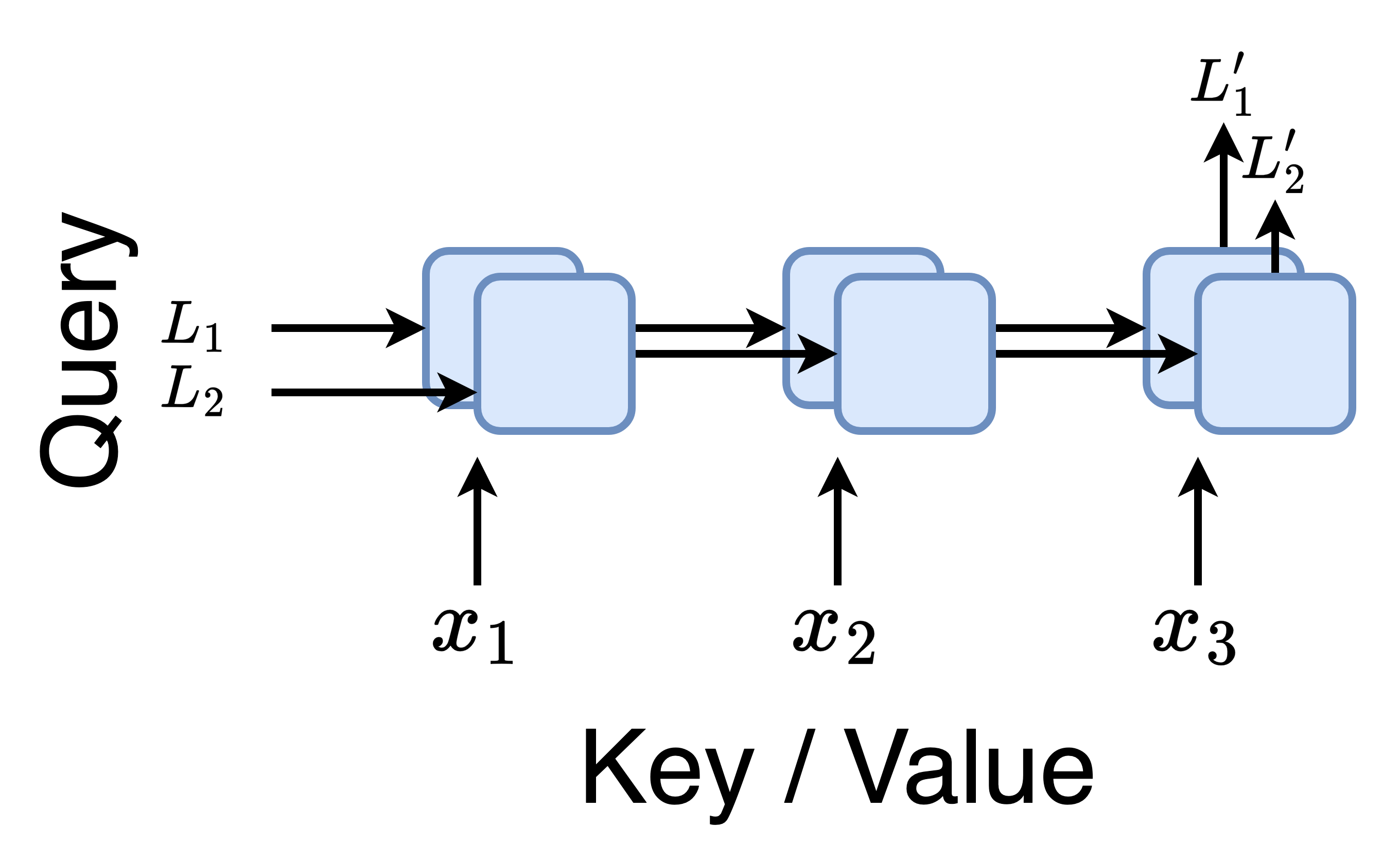}
        \caption{Perceiver's Cross-Attention}
        \label{fig:perceiver}
    \end{subfigure}
    \caption{Attention as a \textit{many-to-one} RNN. The query tokens are the initial hidden states of the RNNs. 
    (a) The conventional method of computing attention only computes its final output. As such, it can be viewed as a method of computing attention's \textit{many-to-one} RNN output. 
    (b) Transformer's self-attention~\citep{vaswani2017attention} uses the input tokens as the initial hidden states.
    (c) Perceiver's cross-attention~\citep{jaegle2021perceiver} uses input-dependent latents as the initial hidden states. 
    }
    \label{fig:Attention-variants}
\end{figure}

Addressing this, we propose an efficient attention-based module capable of leveraging GPU parallelism while being efficiently updateable. 
We begin by first showing in Section \ref{03:attention_many_to_one_RNN} that attention can be viewed as an RNN with the special ability to compute its
\textit{many-to-one} RNN (Figure \ref{fig:many_to_one_AttRNN}) output efficiently. 
Leveraging the RNN formulation of attention, we further show that popular attention-based models such as Transformers (Figure \ref{fig:causal-transformer}) and Perceivers (Figure \ref{fig:perceiver}) can be viewed as RNNs.  
However, unlike traditional RNNs, these models are unable to efficiently update themselves with new tokens, limiting their potential in sequential problem settings where data arrives as a stream. 
Tackling this, we introduce in Section \ref{03:attention_many_to_many_RNN} an efficient method for computing attention as a \textit{many-to-many} RNN based on the parallel prefix scan algorithm. 
Building on this, we introduce in Section \ref{03:Aaren} \textbf{Aaren} (\textbf{[A]}ttention \textbf{[a]}s a \textbf{[re]}current neural \textbf{[n]}etwork), a computationally efficient module that can not only (i) be trained in parallel (like Transformers) but also (ii) be efficiently updated with new tokens at inference time, requiring only constant memory for inferences (like traditional RNNs).

\subsection{Attention as a (many-to-one) RNN}
\label{03:attention_many_to_one_RNN}

\begin{wrapfigure}{R}{0.43\textwidth}
        \centering
    \includegraphics[height=2in]{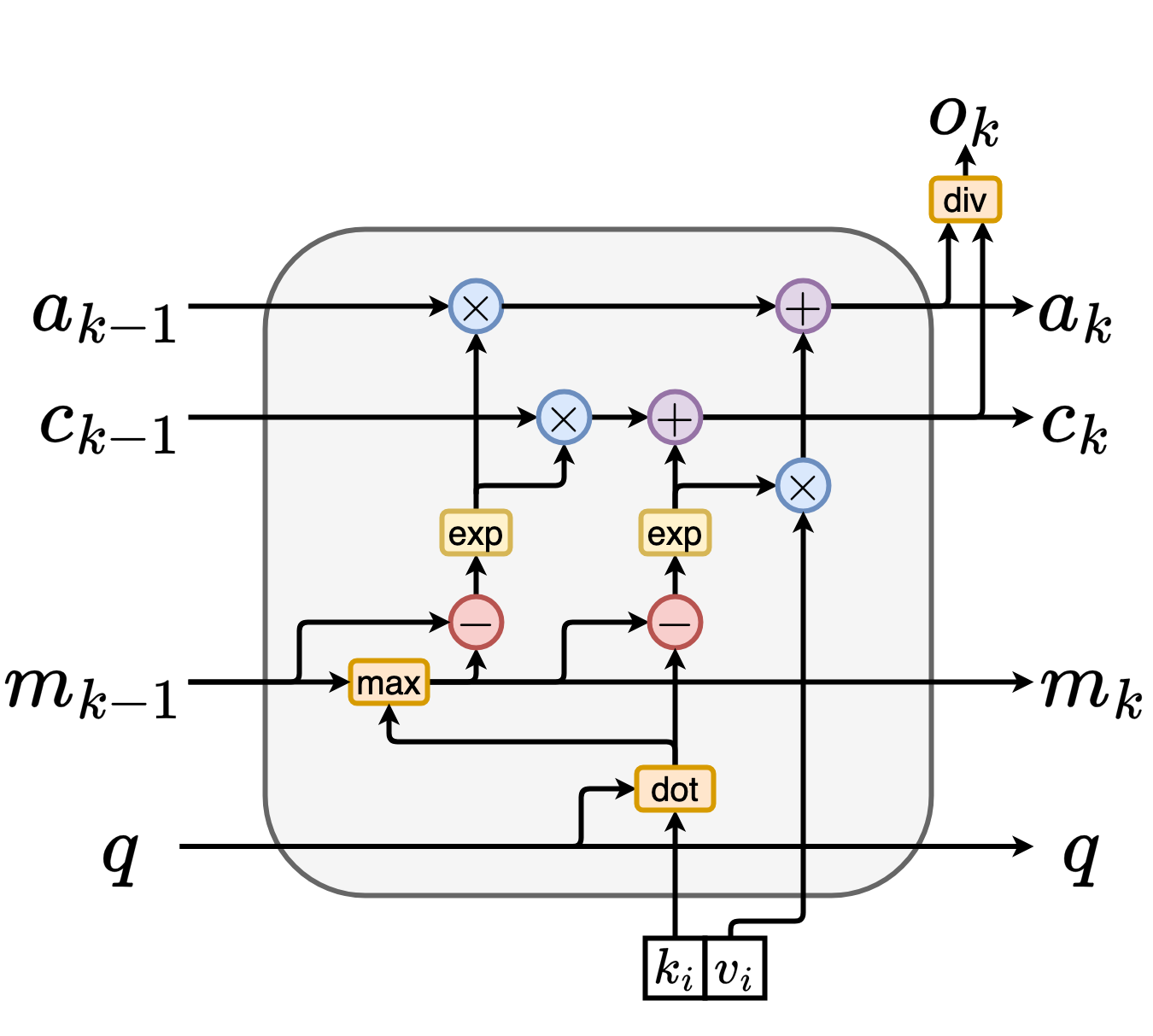}
    \caption{Attention's RNN Cell.
    }
    \label{fig:attention_rnn_block}
\end{wrapfigure}

Attention for a query vector $q$ can be viewed as a function that maps $N$ context tokens $x_{1:N}$ via their keys and values $\{(k_i, v_i)\}_{i=1}^{N}$ to a singular output $o_N = \mathrm{Attention}(q, k_{1:N}, v_{1:N})$. Given $s_{i} = \mathrm{dot}(q, k_i)$, the output $o_N$ can be formulated as:
$$
    o_N =  
    \sum_{i=1}^N \mathrm{softmax}(s)_i v_i = 
    \frac{\sum_{i=1}^N \exp(s_i) v_i}{\sum_{i=1}^N \exp(s_i)} =  \frac{\hat{a}_N}{\hat{c}_N}
$$
where the numerator is $\hat{a}_N = \sum_{i=1}^N \exp(s_i) v_i$ and the denominator is $\hat{c}_N = \sum_{i=1}^N \exp(s_i)$.
Viewing attention as an RNN, we can compute both iteratively as rolling sums $\hat{a}_{k} = \hat{a}_{k-1} + \exp(s_k) v_k$ and $\hat{c}_k = \hat{c}_{k-1} + \exp(s_k)$ for $k = 1, \ldots, N$. However, in practice, this is an unstable implementation, running into numerical issues due to limited precision representations and potentially very small or large exponents (i.e., $\exp(s)$). 
Mitigating this, we re-write the recurrence with a cumulative maximum term\footnote{A common trick for computing softmax numerically stably is to subtract the max term~\citep{goodfellow2016deep}. The cumulative max is an extension of this trick for a recurrent computation of softmax.} $m_k = \mathrm{max}_{i \in \{1, \ldots, k\}} s_i$, computing instead $a_k = \sum_{i=1}^{k} \exp(s_i - m_k) v_i$ and $c_k = \sum_{i=1}^{k} \exp(s_i - m_k)$. 
Notably, the final result is the same $o_N = \frac{\hat{a}_N}{\hat{c}_N} = \frac{a_N}{c_N}$.
$a_k$, $c_k$, and $m_k$ are thus computed recurrently as follows:
\begin{align*}
    a_{k} &= a_{k-1} \exp(m_{k-1} - m_{k}) + v_k \exp(s_k - m_{k}) \\
    c_{k} &= c_{k-1} \exp(m_{k-1} - m_{k}) + \exp(s_k - m_{k}) \\
    m_{k} &= \mathrm{max}(m_{k-1}, s_{k})
\end{align*}
By encapsulating the recurrent computation of $a_k$, $c_k$, and $m_k$ from $a_{k-1}$, $c_{k-1}$, and $m_{k-1}$, we introduce an RNN cell that iteratively computes the output of attention (see Figure \ref{fig:attention_rnn_block}). 
Attention's RNN cell takes as input $(a_{k-1}, c_{k-1}, m_{k-1}, q)$ and computes $(a_k, c_k, m_k, q)$. 
Note that the query vector $q$ is carried over in the RNN cell. 
The initial hidden state of attention's RNN is $(a_0, c_0, m_0, q) = (0, 0, 0, q)$.

\textbf{Methods for computing attention.} By viewing attention as an RNN, we can see that there are different ways to compute attention: (1) recurrently token-by-token (i.e., sequentially) in $O(1)$ memory or (2) in the conventional manner (i.e., in parallel) requiring linear $O(N)$ memory. 
Since attention can be viewed as an RNN, the conventional method of computing attention can also be viewed as an efficient method of computing attention's \textit{many-to-one} RNN output, i.e., the output of an RNN that takes as input multiple context tokens but only outputs a single token at the end of the RNN (see Figure \ref{fig:many_to_one_AttRNN}).
Lastly, instead of fully sequential or fully in parallel, we can also compute attention as (3) an RNN that processes the tokens block-by-block requiring $O(b)$ memory where $b$ is the size of the block. This method, however, is outside the scope of this work. As such, the description of the block-by-block RNN is included in Appendix \ref{appendix:block_by_block_rnn}. 

\textbf{Viewing existing attention-based models as RNNs.} By viewing attention as an RNN, existing attention-based models can also be viewed as variations of RNNs.
For example, Transformers' self-attentions are RNNs (Figure \ref{fig:causal-transformer}) with the context tokens as their initial hidden states.
Perceiver's cross-attentions are RNNs (Figure \ref{fig:perceiver}) with context-dependent latents as their initial hidden states.
By leveraging the RNN formulation of their attention mechanisms, these existing models can compute their output memory efficiently. 

\begin{wrapfigure}{R}{0.43\textwidth}
\begin{minipage}{0.43\textwidth}
    \begin{algorithm}[H]
    \caption{Parallel Prefix Scan (\citet{Hillis1986DataPA}'s variation)}\label{alg:prefix_scan}
    \begin{algorithmic}
    \Require Associative Operator $\oplus$ and $\{x_i\}_{i=1}^{N}$
    \Ensure $\{z_k = \bigoplus_{i=1}^{k} x_i\}_{k=1}^{N}$
    \State $z \leftarrow x$
    \For{$i \leftarrow 1, \ldots, \floor{\log(N)}$}
        \For{$j \leftarrow 0, \ldots, N-1$} \textbf{in parallel}
            \If {$j < 2^i$}
                \State $z'_j \leftarrow z_j$
            \Else
                \State $z'_j \leftarrow z_j \oplus z_{j - 2^i}$
            \EndIf
        \EndFor
        \State $z \leftarrow z'$
    \EndFor
    \end{algorithmic}
    \end{algorithm}
\end{minipage}
\end{wrapfigure}

\textbf{Challenges of viewing attention as an RNN for existing models.} However, when viewing existing attention-based models such as Transformers as RNNs, the models lack important properties common in traditional RNNs such as LSTMs and GRUs. 
Notably, LSTMs and GRUs are capable of efficiently updating themselves with new tokens in only $O(1)$ constant memory and computation, an important feature for sequence modelling where data is received in a stream.  
In contrast, the RNN view of Transformer (see Figure \ref{fig:causal-transformer}) would handle new tokens by adding a new RNN with the new token as its initial state. 
The new RNN processes all preceding tokens, requiring $O(N)$ linear computation in the number of tokens. 
In Perceiver, the latents ($L_i$ in Figure \ref{fig:perceiver}) are input-dependent due to their architecture, meaning that their values change when receiving a new token. Since the initial hidden states (i.e., latents) of their RNNs change, Perceiver would thus require re-computing their RNNs from scratch, requiring $O(NL)$ linear computation in the number of tokens ($N$) and the number of latents ($L$).

\input{core/02_many-to-many}

\subsection{Aaren: Attention as a Recurrent Neural Network}
\label{03:Aaren}

Leveraging the parallelized many-to-many formulation of attention, we propose \textbf{Aaren} (\textbf{[A]}ttention \textbf{[a]}s a \textbf{[re]}current neural \textbf{[n]}etwork). 
The interface of Aaren is the same as a Transformer, mapping $N$ inputs to $N$ outputs whereas the $i$-th output is an aggregate of the $1$st to $i$-th input. 
As such, Aaren is also (1) naturally stackable and (2) capable of computing individual loss terms for each sequence token. 
However, unlike Transformers which use a causal self-attention, Aaren uses the aforementioned method of computing attention as a many-to-many RNN, making it more efficient. 
Aaren functions as follows:
\begin{align*}
    h_1^{(0)}, \ldots, h_N^{(0)} &\leftarrow x_1, \ldots, x_N \\ 
    [h_1^{(j+1)}, \ldots, h_N^{(j+1)}] &\leftarrow \mathrm{Aaren}(q^{(j)}, [h_1^{(j)}, \ldots, h_N^{(j)}]) \\
\end{align*}
Unlike Transformers where the query is one of the input tokens to attention, Aaren's query token $q$ is learned during training via backpropagation. 
In Figure \ref{fig:Stacking-Aaren-Seq-Data}, we include an example of a stacked Aaren model with input context tokens $x_{1:3}$ and outputs $y_{1:3}$. 
Notably, since Aaren leverages the RNN formulation of attention, the stacking of Aarens is also the stacking of RNNs. 
Therefore, Aarens are also able to perform updates with new tokens efficiently, i.e., the iterative computation of $y_k$ only requires constant computation as it relies solely on $h_{k-1}$ and $x_k$. 
Unlike Transformer-based models which (1) require linear memory (when using KV-caching) and (2) \textbf{require} storing all previous tokens, including those in intermediate Transformer layers, Aaren-based models (1) require only constant memory and (2) \textbf{does not require} storing all previous tokens, making Aarens significantly more computationally efficient than Transformers.

%% file: core/02_many-to-many.tex
\subsection{Attention as a (many-to-many) RNN}
\label{03:attention_many_to_many_RNN}

\begin{wrapfigure}{R}{0.43\textwidth}
\vspace{-23pt}
        \centering
        \includegraphics[width=2.3in]{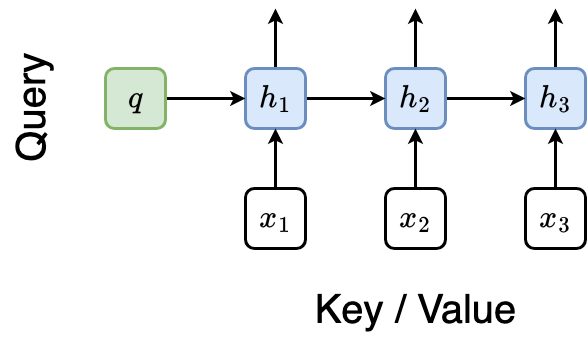}
        \caption{Attention as a many-to-many RNN}
        \label{fig:Aaren}
    \includegraphics[width=2.3in]{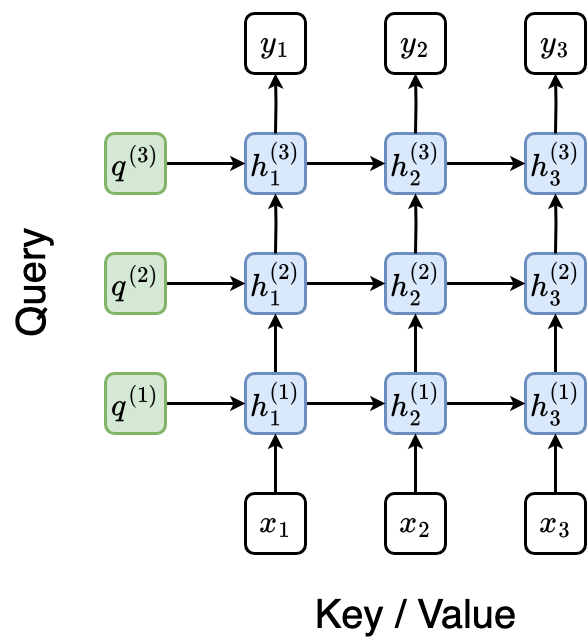}
    \caption{Stacking Aarens for sequence modelling
    }
    \label{fig:Stacking-Aaren-Seq-Data}
\end{wrapfigure}

Addressing these limitations, we propose to develop an attention-based model capable of leveraging the RNN formulation's ability to perform efficient updates. 
To do so, we first introduce an efficient parallelized method of computing attention as a \textit{many-to-many} RNN, i.e., a parallel method to compute $\{o_i = \mathrm{Attention}(q, x_{1:i})\}_{i=1}^{N}$. 
To do so, we leverage the parallel prefix scan algorithm~\citep{blelloch1990prefix} (see Algorithm \ref{alg:prefix_scan}), a parallel computation method for computing $N$ prefix computations from $N$ sequential data points via an associative operator $\oplus$. The algorithm efficiently computes $\{\bigoplus_{i=1}^{k} x_i\}_{k=1}^{N}$ from $\{x_k\}_{k=1}^{N}$.

Recall that $\mathrm{Attention(q, x_{1:k})} = o_k = \frac{a_k}{c_k}$ where $a_k = \sum_{i=1}^{k} \exp(s_i - m_k) v_i$, $c_k = \sum_{i=1}^{k} \exp(s_i - m_k)$, and $m_k = \mathrm{max}_{i \in \{1, \ldots, k\}} s_i$
To compute $\{\mathrm{Attention(q, x_{1:k})}\}_{k=1}^{N}$ efficiently, we can compute $\{a_k\}_{k=1}^{N}$, $\{c_k\}_{k=1}^{N}$, and $\{m_k\}_{k=1}^{N}$ via the parallel scan algorithm and afterwards combine $a_k$ and $c_k$ to compute $\mathrm{Attention(q, x_{1:k})}$. 

To do so, we propose the following associative operator $\oplus$ that acts on $3$-tuples\footnote{See Appendix  \ref{appendix:parallel_scan} for proof of the correctness and associative nature of $\oplus$.} of the form $(\mathtt{m}_A, \mathtt{u}_A, \mathtt{w}_A)$ where $A$ is a set of indices, $\mathtt{m}_A = \mathrm{max}_{i \in A} s_i$, $\mathtt{u}_A = \sum_{i \in A} \exp(s_i - \mathtt{m}_A)$, and $\mathtt{w}_A = \sum_{i \in A} \exp(s_i - \mathtt{m}_A)v_i$. The parallel scan algorithm takes as input $\{(\mathtt{m}_{\{i\}}, \mathtt{u}_{\{i\}}, \mathtt{w}_{\{i\}})\}_{i=1}^{N} = \{(s_i, 1, v_i)\}_{i=1}^{N}$.
The algorithm recursively applies the operator $\oplus$ which works as follows:
$$
(\mathtt{m}_{A}, \mathtt{u}_{A}, \mathtt{w}_{A}) \oplus (\mathtt{m}_{B}, \mathtt{u}_{B}, \mathtt{w}_{B}) = 
(\mathtt{m}_{A \cup B}, \mathtt{u}_{A \cup B}, \mathtt{w}_{A \cup B})
$$
where $\mathtt{m}_{A \cup B} = \mathrm{max}(\mathtt{m}_{A}, \mathtt{m}_{B})$, $\mathtt{u}_{A \cup B} = \mathtt{u}_{A} \exp(\mathtt{m}_{A} - \mathtt{m}_{A \cup B}) + \mathtt{u}_{B} \exp(\mathtt{m}_{B} - \mathtt{m}_{A \cup B})$ and $\mathtt{w}_{A \cup B} = \mathtt{w}_{A} \exp(\mathtt{m}_{A} - \mathtt{m}_{A \cup B}) + \mathtt{w}_{B} \exp(\mathtt{m}_{B} - \mathtt{m}_{A \cup B})$.
Upon completion of applying the operator recursively, the algorithm outputs 
$\{(\mathtt{m}_{\{1, \ldots, k\}}, \mathtt{u}_{\{1, \ldots, k\}}, \mathtt{w}_{\{1, \ldots, k\}})\}_{k=1}^{N} = 
\{(m_k, \sum_{i=1}^{k} \exp(s_i - m_k), \sum_{i=1}^{k} \exp(s_i - m_k) v_i)\}_{k=1}^{N}$. Also known as $\{(m_k, c_k, a_k)\}_{k=1}^{N}$. Combining the last two values of the output tuples, we retrieve $\mathrm{Attention(q, x_{1:k})} = o_k = \frac{a_k}{c_k}$, resulting in an efficient parallelized method for computing attention as a many-to-many RNN (Figure \ref{fig:Aaren}).

%% file: core/03_experiments.tex
\section{Experiments} \label{sec:experiments}

Our objective in the experiments is to compare Aarens with Transformers in terms of (1) performance and (2) resources required (time and memory). 
To perform a comprehensive comparison, we evaluate across four problem settings: reinforcement learning, event forecasting, time series forecasting, and time series classification.

\textbf{Datasets.} 
In total, we evaluate Aarens and Transformers on $38$ datasets, the majority of which are real-world datasets. The datasets are split between problem settings as follows: $12$ reinforcement learning datasets, $8$ event forecasting datasets, $8$ time series forecasting datasets, and $10$ time series classification datasets. 
For each dataset, the models are evaluated with $5$ seeds.
Due to space limitations, we refer the reader to Appendix \ref{appendix:dataset} descriptions of individual datasets. 

\textbf{Models.} 
To compare Aarens directly with Transformers, we replace the Transformers with Aarens in domain-specialized Transformer models. 
For reinforcement learning, we performed the comparison on Decision Transformer~\citep{chen2021decision}. 
For event forecasting, we performed the comparison Transformer Hawkes Process~\citep{zuo2020transformer, bae2023meta}.
For time series forecasting, we performed the comparison on a Transformer with input normalization following \citet{liu2022non};
For time series classification, we performed the comparison on a vanilla causal transformer following the library by \citet{wu2023timesnet}.

\textbf{Implementation Details.} 
The experiments are run using the datasets and transformer implementations of popular repositories. 
More specifically, the reinforcement learning experiments with Decision Transformer were run on the code by \citet{minimal_decision_transformer}.
The time series forecasting and time series classification experiments were run on the Time Series Library repository by \citet{wu2023timesnet}.
The event forecasting experiments were run on the code by \citet{bae2023meta}. 
As Transformers and Aarens share the same interface and are both attention-based methods, they share the same set of hyperparameters. 
For fairness, the same hyperparameters are used for both Transformers and Aarens. 
Due to space limitations, specific hyperparameter details are included in Appendix \ref{appendix:implementation_details}\footnote{The code will be released alongside the camera-ready.}.

\subsection{Reinforcement Learning}

In these experiments, we compare Aarens with Transformers on reinforcement learning (RL). 
In RL, the model's objective during training is to learn a policy that learns from feedback/rewards obtained through trial and error while interacting in an environment. 
As such, RL is popular in interactive settings such as robotics, recommendation engines, and traffic control. 
For these experiments, we consider Decision Transformers~\citep{chen2021decision}, a popular method for training a policy in an offline manner on datasets of environment interactions. 
At test time, Decision Transformers are evaluated in an online manner. 

We evaluate on
popular locomotion robotics environments from the D4RL benchmark~\citep{fu2020d4rl}: HalfCheetah, Ant, Hopper, and Walker. 
For each of the locomotion environments, we compare the models trained on three different kinds of datasets: Medium, Medium-Replay, and Medium-Expert. 
Each of these datasets consists of $1$ million timesteps generated by a policy. 
In total, we compare model performances across $4 \times 3 = 12$ RL datasets.
We refer the reader to Appendix \ref{appendix:dataset:rl} for more details regarding the individual tasks.
The results in Table \ref{03:rl:table:results} show that Aarens achieve competitive performance with Transformers across all twelve datasets and four environments. 
However, unlike Transformers, Aarens are able to efficiently process new environmental interactions in constant computation due to also being an RNN, making it more suitable for reinforcement learning.

\begin{table}[]

\resizebox{\textwidth}{!}{%
\bgroup
\def\arraystretch{1.25}
\begin{tabular}{c|ccc|ccc}
\hline
\multicolumn{7}{c}{\large \textbf{Reinforcement Learning (Score $\uparrow$)}}                                                                                                                  \\ \hline
\multicolumn{1}{c|}{\multirow{2}{*}{Dataset}} & \multicolumn{3}{c|}{HalfCheetah}                                                                         & \multicolumn{3}{c}{Ant}                                                             \\ \cline{2-7} 
\multicolumn{1}{c|}{}                         & Medium                    & Med-Replay                & \multicolumn{1}{c|}{Med-Expert}                  & Medium                    & Med-Replay                & Med-Expert                  \\ \hline
\multicolumn{1}{c|}{Transformer}              & $\mathbf{41.88 \pm 1.47}$ & $\mathbf{36.57 \pm 1.40}$ & \multicolumn{1}{c|}{$\mathbf{75.98 \pm 6.34}$}   & $\mathbf{94.25 \pm 8.62}$ & $\mathbf{89.39 \pm 4.96}$ & $\mathbf{125.47 \pm 10.99}$ \\
\multicolumn{1}{c|}{Aaren}                    & $\mathbf{42.16 \pm 1.89}$ & $\mathbf{37.91 \pm 1.94}$ & \multicolumn{1}{c|}{$\mathbf{75.74 \pm 15.13}$}  & $\mathbf{93.29 \pm 4.04}$ & $\mathbf{85.53 \pm 6.57}$ & $\mathbf{119.72 \pm 12.63}$ \\ \hline
\multicolumn{1}{c|}{\multirow{2}{*}{Dataset}} & \multicolumn{3}{c|}{Hopper}                                                                              & \multicolumn{3}{c}{Walker}                                                          \\ \cline{2-7} 
\multicolumn{1}{c|}{}                         & Medium                    & Med-Replay                & \multicolumn{1}{c|}{Med-Expert}                  & Medium                    & Med-Replay                & Med-Expert                  \\ \hline
\multicolumn{1}{c|}{Transformer}              & $\mathbf{80.18 \pm 5.85}$ & $\mathbf{79.73 \pm 7.64}$ & \multicolumn{1}{c|}{$\mathbf{98.82 \pm 10.33}$}  & $\mathbf{77.84 \pm 3.81}$ & $\mathbf{72.36 \pm 5.63}$ & $\mathbf{109.66 \pm 0.45}$  \\
\multicolumn{1}{c|}{Aaren}                    & $\mathbf{80.86 \pm 4.77}$ & $\mathbf{77.87 \pm 5.68}$ & \multicolumn{1}{c|}{$\mathbf{103.89 \pm 11.89}$} & $\mathbf{74.44 \pm 5.16}$ & $\mathbf{71.44 \pm 6.55}$ & $\mathbf{110.51 \pm 1.30}$ 

\end{tabular}
\egroup
}
\caption{Reinforcement Learning Results. Measurement of the D4RL score (higher is better)~\citep{fu2020d4rl}. 
 The \textbf{bolded} results indicate the best-performing method.  
}
\label{03:rl:table:results}
\end{table}

\subsection{Event Forecasting}

In these experiments, we compare Aarens with Transformers on event forecasting (EF). 
In EF, the model is given a sequence of irregularly spaced discrete events in time and models the probability distribution of the next event time and its mark (i.e., event label/class). 
EF is popular in many real-world settings such as finance (e.g., transactions), healthcare (e.g., patient observations), and e-commerce (e.g., purchases) where user or system actions occur at irregularly spaced times.
To perform the comparison, we replace the Transformers in Transformers Hawkes Process~\citep{zuo2020transformer} with Aarens. 
Following \citet{bae2023meta}, a mixture of log-normal distributions is used to model the probability distribution of the next event time.
For this setting, we consider $8$ popular benchmarking datasets for next event forecasting~\citep{zhang2020self, zuo2020transformer, bae2023meta}: MIMIC, Wiki, Reddit, Mooc, StackOverflow, Sin, Uber, and Taxi. 
$7$ out of the $8$ are real-world datasets whereas only Sin is a synthetic dataset.
$3$ out of the $8$ datasets (Sin, Uber, and Taxi) do not include marks/labels. 
We refer the reader to Appendix \ref{appendix:dataset:event_forecasting} for details regarding individual datasets.
The results in Table \ref{table:tpp} show that Aarens performed comparably with Transformers across all datasets. 
Aaren's ability to efficiently process new inputs is a particularly useful feature in event forecasting settings where events arrive in an irregular stream.

\begin{table}[ht]
\centering
\resizebox{\textwidth}{!}{%
\bgroup
\def\arraystretch{1.25}
\begin{tabular}{c|cc|cc|cc}
 \hline
\multicolumn{7}{c}{\large \textbf{Event Forecasting}}                                                                                                                                              \\ \hline

\large Metric      & \multicolumn{2}{c|}{\large \textbf{NLL} ($\downarrow$)}                  & \multicolumn{2}{c|}{\large \textbf{RMSE} ($\downarrow$)}               & \multicolumn{2}{c}{\large \textbf{Acc} ($\uparrow$)}               \\ \hline
\multicolumn{1}{c|}{Dataset}                      & MIMIC                    & \multicolumn{1}{c|}{Wiki}                      & MIMIC                     & \multicolumn{1}{c|}{Wiki}                      & MIMIC                              & Wiki                              \\ \hline
\multicolumn{1}{c|}{Transformer}                  & $\mathbf{1.22 \pm 0.08}$ & \multicolumn{1}{c|}{$\mathbf{9.66 \pm 0.98}$}  & $\mathbf{1.60 \pm 0.28}$  & \multicolumn{1}{c|}{$\mathbf{0.28 \pm 0.04}$}  & $\mathbf{84.07 \pm 1.46}$          & $\mathbf{23.60 \pm 2.66}$         \\
\multicolumn{1}{c|}{Aaren}                        & $\mathbf{1.21 \pm 0.06}$ & \multicolumn{1}{c|}{$\mathbf{8.98 \pm 1.03}$}  & $\mathbf{1.56 \pm 0.32}$  & \multicolumn{1}{c|}{$\mathbf{0.22 \pm 0.05}$}  & $\mathbf{84.53 \pm 0.66}$          & $\mathbf{21.26 \pm 1.29}$         \\ \hline
\multicolumn{1}{c|}{Dataset}                      & Reddit                   & \multicolumn{1}{c|}{Mooc}                      & Reddit                    & \multicolumn{1}{c|}{Mooc}                      & Reddit                             & Mooc                              \\ \hline
\multicolumn{1}{c|}{Transformer}                  & $\mathbf{0.40 \pm 0.29}$ & \multicolumn{1}{c|}{$\mathbf{-0.22 \pm 0.57}$} & $\mathbf{0.23 \pm 0.03}$  & \multicolumn{1}{c|}{$\mathbf{0.20 \pm 0.04}$}  & $\mathbf{60.68 \pm 1.62}$          & $\mathbf{37.79 \pm 0.42}$         \\
\multicolumn{1}{c|}{Aaren}                        & $\mathbf{0.31 \pm 0.30}$ & \multicolumn{1}{c|}{$\mathbf{0.25 \pm 1.61}$}  & $\mathbf{0.30 \pm 0.04}$  & \multicolumn{1}{c|}{$\mathbf{0.41 \pm 0.28}$}  & $\mathbf{62.34 \pm 0.40}$          & $\mathbf{36.69 \pm 1.48}$         \\ \hline
\multicolumn{1}{c|}{Dataset}                      & StackOverflow            & \multicolumn{1}{c|}{Sin}                       & StackOverflow             & \multicolumn{1}{c|}{Sin}                       & StackOverflow                      &                                   \\ \hline
\multicolumn{1}{c|}{Transformer}                  & $\mathbf{2.92 \pm 0.04}$ & \multicolumn{1}{c|}{$\mathbf{0.68 \pm 0.05}$}  & $\mathbf{1.44 \pm 0.08}$  & \multicolumn{1}{c|}{$\mathbf{1.75 \pm 0.09}$}  & $\mathbf{46.44 \pm 0.08}$          &                                   \\
\multicolumn{1}{c|}{Aaren}                        & $\mathbf{2.91 \pm 0.02}$ & \multicolumn{1}{c|}{$\mathbf{0.78 \pm 0.13}$}  & $\mathbf{1.27 \pm 0.17}$  & \multicolumn{1}{c|}{$\mathbf{2.03 \pm 0.25}$}  & $\mathbf{46.34 \pm 0.21}$          &                                   \\ \hline
\multicolumn{1}{c|}{Dataset}                      & Uber                     & \multicolumn{1}{c|}{Taxi}                      & Uber                      & \multicolumn{1}{c|}{Taxi}                      &                                    &                                   \\ \hline
\multicolumn{1}{c|}{Transformer}                  & $\mathbf{3.33 \pm 0.14}$ & \multicolumn{1}{c|}{$\mathbf{2.01 \pm 0.17}$}  & $73.63 \pm 5.73$          & \multicolumn{1}{c|}{$\mathbf{10.34 \pm 0.32}$} &                                    &                                   \\
\multicolumn{1}{c|}{Aaren}                        & $\mathbf{3.48 \pm 0.10}$ & \multicolumn{1}{c|}{$2.33 \pm 0.12$}           & $\mathbf{54.61 \pm 5.40}$ & \multicolumn{1}{c|}{$\mathbf{10.01 \pm 0.52}$} &                                    &                                  

\end{tabular}
\egroup
}

\caption{Event Forecasting Results. Sin, Uber, and Taxi datasets do not include marks/labels. 
}
\label{table:tpp}
\end{table}

\subsection{Time Series Forecasting}

In these experiments, we compared Aarens with Transformers on time series forecasting (TSF).
In TSF, the model is given a series of observations of temporally continuous signals. 
The objective of the model is to predict $T$ future values of the series. 
TSF models are commonly used in a wide range of domains, including those related to climate (e.g., weather), energy (e.g., supply and demand), and economics (e.g., stock prices).
To perform the comparison, we consider a causally masked Transformer with input normalization following \citet{liu2022non}. 
For this setting, we consider $8$ real-world datasets used in prior works: Weather, Exchange, Traffic, ECL, ETTh1, ETTh2, ETTm1, and ETTm2. 
For details regarding the individual datasets, we refer the reader to Appendix \ref{appendix:dataset:tsf}.
Following \citet{wu2023timesnet}, the models are evaluated with $T \in \{96, 192, 336, 720\}$ given an input length of $96$. 
Due to space limitations, Table \ref{table:tsf} only includes results for $T=192$.
We refer the reader to Table \ref{appendix:table:tsf:full_results} in Appendix \ref{appendix:experiments} for the full results.
The results in Table \ref{table:tsf} show that Aarens perform comparably with Transformers across all datasets. 
However, unlike Transformers, Aarens efficiently processes the time series data, making it more suitable for time series-related domains.

\begin{table}[]

\resizebox{\textwidth}{!}{%
\bgroup
\def\arraystretch{1.25}
\begin{tabular}{ccccccc}
\hline
\multicolumn{7}{c}{\large \textbf{Time Series Forecasting}}                                                                                                                                                                    \\ \hline
\multicolumn{1}{c|}{\large Metric}      & \multicolumn{3}{c|}{\large \textbf{MSE ($\downarrow$)}}                                                    & \multicolumn{3}{c}{\large \textbf{MAE ($\downarrow$)}}                                \\ \hline
\multicolumn{1}{c|}{Dataset}     & Weather                  & Exchange                 & \multicolumn{1}{c|}{Traffic}                  & Weather                  & Exchange                 & Traffic                  \\ \hline
\multicolumn{1}{c|}{Transformer} & $\mathbf{0.24 \pm 0.01}$ & $\mathbf{0.24 \pm 0.02}$ & \multicolumn{1}{c|}{$\mathbf{0.63 \pm 0.01}$} & $\mathbf{0.28 \pm 0.00}$ & $\mathbf{0.34 \pm 0.01}$ & $\mathbf{0.34 \pm 0.00}$ \\
\multicolumn{1}{c|}{Aaren}       & $\mathbf{0.25 \pm 0.01}$ & $\mathbf{0.25 \pm 0.03}$ & \multicolumn{1}{c|}{$\mathbf{0.64 \pm 0.01}$} & $\mathbf{0.28 \pm 0.00}$ & $\mathbf{0.33 \pm 0.02}$ & $0.35 \pm 0.00$ \\ \hline
\multicolumn{1}{c|}{Dataset}     & ETTh1                    & ETTm1                    & \multicolumn{1}{c|}{ECL}                      & ETTh1                    & ETTm1                    & ECL                      \\ \hline
\multicolumn{1}{c|}{Transformer} & $\mathbf{0.64 \pm 0.05}$ & $\mathbf{0.52 \pm 0.05}$ & \multicolumn{1}{c|}{$\mathbf{0.39 \pm 0.03}$} & $\mathbf{0.57 \pm 0.02}$ & $\mathbf{0.47 \pm 0.01}$ & $\mathbf{0.48 \pm 0.02}$ \\
\multicolumn{1}{c|}{Aaren}       & $\mathbf{0.59 \pm 0.03}$ & $\mathbf{0.51 \pm 0.03}$ & \multicolumn{1}{c|}{$\mathbf{0.37 \pm 0.02}$} & $\mathbf{0.55 \pm 0.01}$ & $\mathbf{0.47 \pm 0.01}$ & $\mathbf{0.45 \pm 0.01}$ \\ \hline
\multicolumn{1}{c|}{Dataset}     & ETTh1                    & ETTm1                    & \multicolumn{1}{c|}{}                         & ETTh2                    & ETTm2                    &                       \\ \hline
\multicolumn{1}{c|}{Transformer} & $\mathbf{0.50 \pm 0.03}$ & $\mathbf{0.38 \pm 0.02}$ & \multicolumn{1}{c|}{}                         & $\mathbf{0.46 \pm 0.01}$ & $\mathbf{0.37 \pm 0.01}$ &                       \\
\multicolumn{1}{c|}{Aaren}       & $\mathbf{0.49 \pm 0.03}$ & $\mathbf{0.34 \pm 0.04}$ & \multicolumn{1}{c|}{}                         & $\mathbf{0.48 \pm 0.02}$ & $\mathbf{0.39 \pm 0.02}$ &                      
\end{tabular}
\egroup
}
\caption{Time Series Forecasting Results. 
Following \citet{wu2023timesnet}, the models are evaluated with $T \in \{96, 192, 336, 720\}$ given an input length of $96$. 
Due to space limitations, this table only includes results for $T=192$.
We refer the reader to Appendix \ref{appendix:experiments} (Table \ref{appendix:table:tsf:full_results}) for the full results.
}
\label{table:tsf}
\end{table}

\subsection{Time Series Classification}

In these experiments, we compared Aarens with Transformers on time series classification (TSC).
In TSC, the model's objective is to predict the label of a time series.
This setting is common in many important applications such as pattern recognition (e.g., electrocardiograms), anomaly detection (e.g., bank fraud), or failure prediction (e.g., power grid fluctuations)~\citep{dinger2022what}. 
For this setting, we consider $10$ real-world popular datasets from the UEA time series classification Archive~\citep{bagnall2018uea}: EthanolConcentration, FaceDetection, Handwriting, Heartbeat, JapaneseVowels, PEMS-SF, SelfRegulationSCP1, SelfRegulationSCP2 ArabicDigits, and UWaveGesture.
For details regarding the individual datasets, we refer the reader to Appendix \ref{appendix:dataset:tsc}.
In Table \ref{table:tsc}, we see that Aarens perform comparably with Transformers across all datasets. 

\begin{table}[]
\resizebox{\textwidth}{!}{%
\bgroup
\def\arraystretch{1.25}
\begin{tabular}{c|ccccc}
\hline
\multicolumn{6}{c}{\large \textbf{Time Series Classification (Acc $\uparrow$)}}                                                                            \\ \hline
Dataset     & EthanolConc.              & FaceDetection             & Handwriting               & Heartbeat                 & Jap. Vowels               \\ \hline
Transformer & $\mathbf{29.89 \pm 1.63}$ & $\mathbf{69.23 \pm 0.52}$ & $\mathbf{26.54 \pm 2.25}$ & $\mathbf{74.05 \pm 1.21}$ & $\mathbf{96.38 \pm 0.91}$ \\
Aaren       & $\mathbf{29.58 \pm 2.30}$ & $\mathbf{69.06 \pm 0.61}$ & $\mathbf{27.39 \pm 1.46}$ & $\mathbf{74.15 \pm 0.77}$ & $\mathbf{96.65 \pm 0.75}$ \\ \hline
Dataset     & PEMS-SF                   & SelfReg. SCP1             & SelfReg. SCP2             & ArabicDigits              & UWaveGesture              \\ \hline
Transformer & $\mathbf{78.73 \pm 2.06}$ & $\mathbf{88.81 \pm 0.92}$ & $\mathbf{52.89 \pm 2.47}$ & $\mathbf{98.89 \pm 0.57}$ & $\mathbf{79.81 \pm 1.51}$ \\
Aaren       & $\mathbf{81.85 \pm 2.60}$ & $\mathbf{89.42 \pm 1.85}$ & $\mathbf{54.22 \pm 1.50}$ & $\mathbf{98.68 \pm 0.20}$ & $\mathbf{82.00 \pm 1.93}$ \\ 
\end{tabular}
\egroup
}
\caption{Time Series Classification Results. 
}
\label{table:tsc}
\end{table}

\subsection{Analyses}

In these experiments, we compare Aarens with Transformers in terms of the resources required. To do so, we use the code by \citet{minimal_decision_transformer}.
For Transformers, we use KV-caching to improve their efficiency.

\begin{figure}[]
\begin{subfigure}{.5\textwidth}
  \centering
  \includegraphics[width=.95\linewidth]{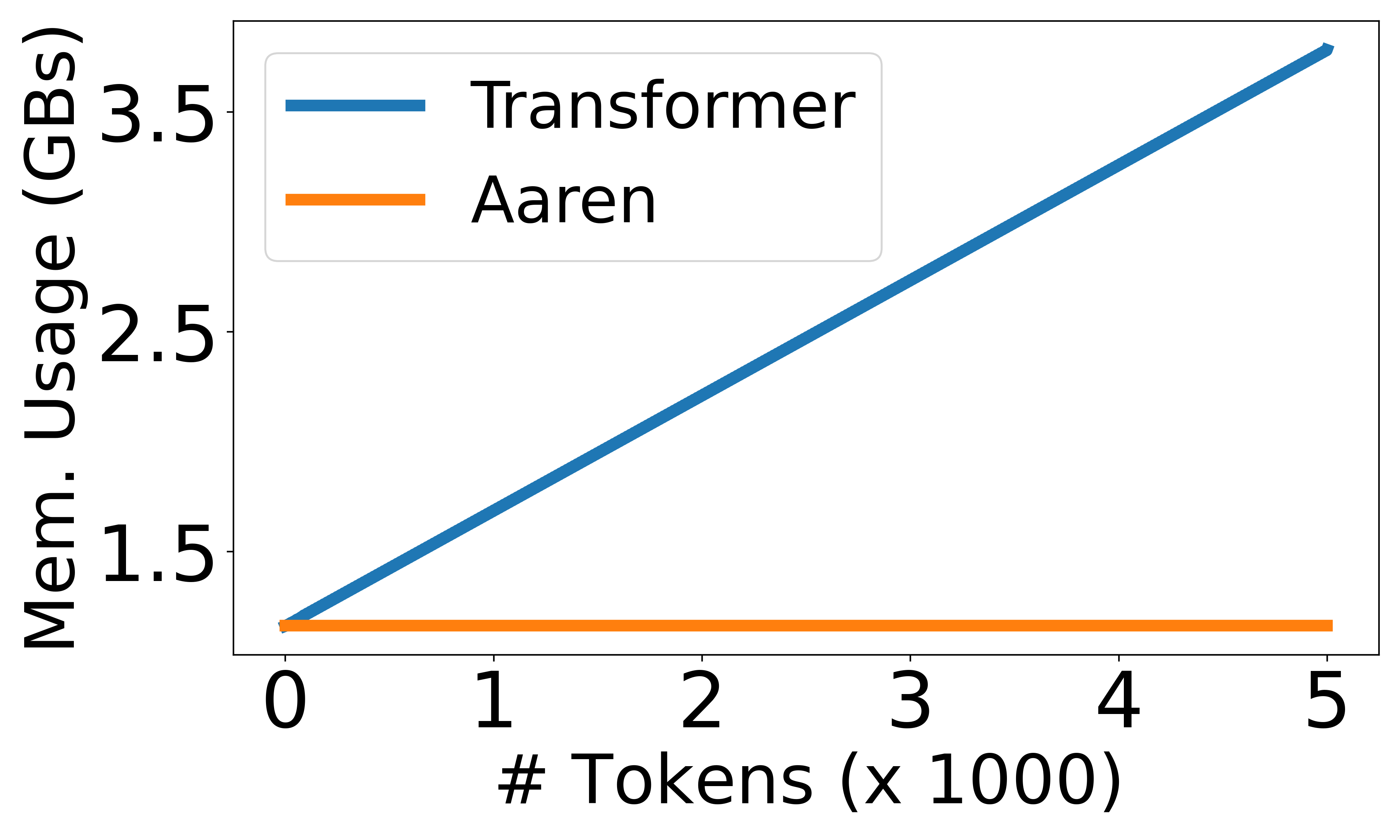}
\end{subfigure}
\begin{subfigure}{.5\textwidth}
  \centering
  \includegraphics[width=.95\linewidth]{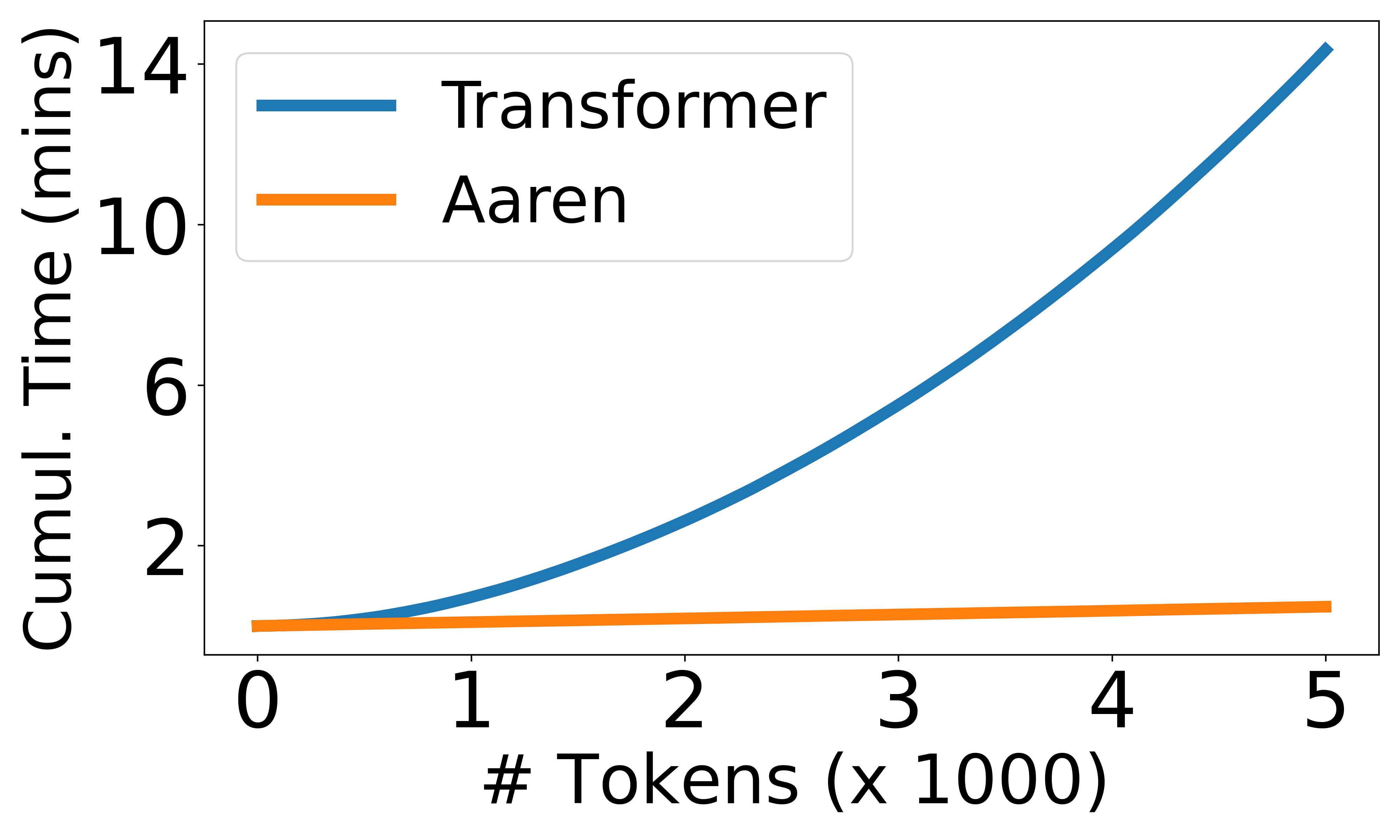}
\end{subfigure}%
\caption{Computational Resources Plots comparing Aarens and Transformers (using KV-caching) when processing a sequence of tokens. (Left) Memory Usage Comparison. (Right) Cumulative Time Comparison. 
}
\label{fig:analyses}
\end{figure}

\textbf{Memory Complexity:} In Figure \ref{fig:analyses} (left), we compare the memory usage of Aarens and Transformers (using KV-caching) at inference time. We see that the memory usage of Transformers grow linearly with the KV-caching technique. In contrast, Aarens only use constant memory regardless of the number of tokens, making it significantly more efficient.

\textbf{Time Complexity:} In Figure \ref{fig:analyses} (right), we compare the cumulative time needed by Aarens and Transformers (using KV-caching) for sequentially processing a sequence of tokens. 
In the case of Transformers, the cumulative amount of computation is quadratic in the number of tokens, i.e., $O(1 + 2 + \ldots + N) = O(N^2)$. 
In contrast, for Aaren, the cumulative amount of computation is linear.
In the figure, we see a similar result in the cumulative time needed by the models.
Specifically, the cumulative time needed by Transformers grows quadratically while that of Aaren grows linearly.

\textbf{Number of Parameters:} Due to learning the initial hidden state $q$, Aaren modules require slightly more parameters than Transformer modules. 
However, the difference is marginal due to $q$ being only a vector. 
Measuring this empirically in comparable models, we found that Transformers used $3,152,384$ parameters. 
In contrast, the equivalent Aarens used  $3,152,896$ parameters, representing only a marginal $\sim0.016\%$ parameter increase -- a minor trade-off for the significant gains in memory and time complexities.

%% file: core/04_relatedwork.tex
\section{Related Work} \label{sec:related_work}

Closest to \textbf{Aaren} are approximations of attention such as those by RWKV~\citep{peng2023rwkv}, RetNet~\citep{sun2023retentive}, and Linear Transformer~\citep{katharopoulos2020transformers}. These models proposed linearizations of the standard softmax-based attention that allow them to be formulated as an RNN. 
However, in doing so, these models also encode an exponential factor that biases tokens based on their timestamp, limiting their potential applications. 
In contrast, \textbf{Aaren} leverages an exact re-formulation of softmax attention as an RNN, allowing the model itself to compute the weight of each token.

\citet{feng2023memory} showed attention can be computed recurrently, using it to compress set-based inputs. \citet{rabe2022self} introduced a recurrent formulation of attention, showing that self-attention can be computed efficiently.
\citet{katharopoulos2020transformers} showed that Transformers with a causal mask can be viewed as an RNN. 
In contrast, we (1) show a more general result whereas any attention model can be viewed as an RNN. Furthermore, we (2) introduce \textbf{Aaren}, a new attention formulation based on parallel prefix sums, that achieves competitive results with that of Transformers while being more efficient.

The problem of computing prefix scans/sums has been well studied with various efficient parallelized algorithms proposed for computing them.
Since \textbf{Aaren} only requires the output of the prefix scan, any efficient algorithm for computing it can be used.
In this work, we outlined the method by \citet{Hillis1986DataPA}. 
This method is time efficient for parallel computation, requiring $\log_2(N)$ sequential steps and $\mathcal{O}(N\log(N))$ overall computation. 
In contrast, the method by \citet{ladner1980parallel} use mores sequential steps (specifically, $2\log_2(N) - 2$) but only performs $\mathcal{O}(N)$ overall computation.
For a more in-depth introduction to parallel prefix sums algorithms, we refer the reader to the following work by \citet{blelloch1990prefix}. 

In this work, we applied Transformers to a subset of applications.
For a broad overview of the applications of Transformers, we refer the reader to the following survey by \citet{islam2023comprehensive}.
For an overview of different transformer models applied to the specific settings considered in this paper, we refer the reader to the following surveys (1) on transformers in reinforcement learning by \citet{li2023survey} and (2) on transformers in event forecasting, time series forecasting, time series classification, and more by \citet{wen2022transformers}.

%% file: core/05_conclusion.tex
\section{Conclusion} \label{sec:conclusion}

In this work, we showed that attention can be formulated as an RNN whereas the conventional way of computing attention is a parallelized method of computing its \textit{many-to-one} RNN output. 
Building on the RNN formulation, we showed that existing attention-based models can be formulated as RNNs. 
However, unlike traditional RNNs such as LSTMs and GRUs, these methods cannot be updated efficiently with new tokens.
Addressing this, we introduced a new parallelized method of computing attention's \textit{many-to-many} RNN output based on the parallel prefix scan algorithm.
Building on the new attention formulation, we introduced \textbf{Aaren}, a new module that can not only (i) be trained in parallel (like Transformers) but also (ii) be efficiently updated at inference time, thereby requiring only constant memory (like RNNs). 
Empirically, we showed that Aarens achieve performance competitive with Transformers on $38$ datasets spread across four sequential data settings: reinforcement learning, event forecasting, time series classification, and time series forecasting. 
Finally, we empirically show that Aarens are significantly more time and memory-efficient than Transformers.

%% file: core/99_appendix.tex
\section*{Appendix}
\input{appendix/99_01_block-by-block}
\input{appendix/99_02_associative}

\input{appendix/99_03_dataset_description}

\input{appendix/99_04_experiments}

\input{appendix/99_05_implementation-details}

\input{appendix/99_06_compute-resources}

\input{appendix/99_07_limitations_broader-impact}

%% file: appendix/99_01_block-by-block.tex
\section{Block-by-block method of computing attention as a (many-to-one) RNN. }
\label{appendix:block_by_block_rnn}

In the main paper, we mentioned there are multiple ways to compute attention: (1) the conventional manner (i.e., in parallel) requiring $O(N)$ linear memory and (2) recurrently token-by-token as an RNN, requiring only $O(1)$ constant memory. 
In between these two methods, there is a third method which computes attention (3) in a block-by-block manner as follows:
\begin{align*}
    a_{i+b} &= a_{i} \exp(m_{i} - m_{i+b}) + \sum_{j=i}^{i+b} v_j \exp(s_j - m_{i+b}) \\
    c_{i+b} &= c_{i} \exp(m_{i} - m_{i+b}) + \sum_{j=i}^{i+b}\exp(s_j - m_{i+b}) \\
    m_{i+b} &= \mathrm{max}(m_{i}, s_{i+1}, \ldots, s_{i+b})
\end{align*}
where $b$ is a block size that controls the rate at which data is processed. 
Computing attention in this manner requires $O(b)$ memory where $b$ is a hyperparameter. 
By setting $b=1$, we retrieve the RNN formulation that processes the tokens one-by-one, i.e., computing $a_i$, $c_i$, $m_i$ using the previous time step $a_{i-1}$, $c_{i-1}$, $m_{i-1}$. 

%% file: appendix/99_02_associative.tex
\section{Parallel Prefix Scan's Operator $\oplus$}
\label{appendix:parallel_scan}

In this work, we proposed a formulation of Attention as a \textit{many-to-many} RNN based on the parallel prefix scan algorithm. 
To do so, we defined an associative operator $\oplus$ that is recursively applied during the algorithm. 
In this section, we show the correctness of the operator and prove that it satisfies the associative property. 

The operator $\oplus$ acts on $3$-tuples of the form $(\mathtt{m}_A, \mathtt{u}_A, \mathtt{w}_A)$ where $A$ is a set of indices, $\mathtt{m}_A = \mathrm{max}_{i \in A} s_i$, $\mathtt{u}_A = \sum_{i \in A} \exp(s_i - \mathtt{m}_A)$, and $\mathtt{w}_A = \sum_{i \in A} \exp(s_i - \mathtt{m}_A)v_i$. 
The operator $\oplus$ functions as follows:
$$
(\mathtt{m}_{A}, \mathtt{u}_{A}, \mathtt{w}_{A}) \oplus (\mathtt{m}_{B}, \mathtt{u}_{B}, \mathtt{w}_{B}) = 
(\mathtt{m}_{A \cup B}, \mathtt{u}_{A \cup B}, \mathtt{w}_{A \cup B})
$$
where:
\begin{align*}
    \mathtt{m}_{A \cup B} &= \mathrm{max}(\mathtt{m}_{A}, \mathtt{m}_{B}) \\
    \mathtt{u}_{A \cup B} &= \mathtt{u}_{A} \exp(\mathtt{m}_{A} - \mathtt{m}_{A \cup B}) + \mathtt{u}_{B} \exp(\mathtt{m}_{B} - \mathtt{m}_{A \cup B}) \\
    \mathtt{w}_{A \cup B} &= \mathtt{w}_{A} \exp(\mathtt{m}_{A} - \mathtt{m}_{A \cup B}) + \mathtt{w}_{B} \exp(\mathtt{m}_{B} - \mathtt{m}_{A \cup B})
\end{align*}

\subsection{Correctness Proof}
\label{appendix:parallel_scan:correctness}

To confirm the correctness of the operator, we need to show:
$\mathtt{m}_{A \cup B} = \mathrm{max}_{i \in A \cup B} s_i$, $\mathtt{u}_{A \cup B} = \sum_{i \in A \cup B} \exp(s_i - \mathtt{m}_{A \cup B})$, and $\mathtt{w}_{A \cup B} = \sum_{i \in A \cup B} \exp(s_i - \mathtt{m}_{A \cup B})v_i$.

Expanding the computation of $\mathtt{m}_{A \cup B}$:
\begin{align*}
    \mathtt{m}_{A \cup B} &= \mathrm{max}(\mathtt{m}_{A}, \mathtt{m}_{B}) \\
    &= \mathrm{max}(\mathrm{max}_{i \in A} s_i, \mathrm{max}_{i \in B} s_i) \\
    &= \mathrm{max}_{i \in A \cup B} s_i
\end{align*}
Expanding the computation of $\mathtt{u}_{A \cup B}$:
\begin{align*}
    \mathtt{u}_{A \cup B} &= \mathtt{u}_{A} \exp(\mathtt{m}_{A} - \mathtt{m}_{A \cup B}) + \mathtt{u}_{B} \exp(\mathtt{m}_{B} - \mathtt{m}_{A \cup B}) \\
    &= \left[\sum_{i \in A} \exp(s_i - \mathtt{m}_A) \right]  \exp(\mathtt{m}_{A} - \mathtt{m}_{A \cup B}) + \left[\sum_{i \in B} \exp(s_i - \mathtt{m}_B) \right] \exp(\mathtt{m}_{B} - \mathtt{m}_{A \cup B}) \\
    &= \sum_{i \in A} \exp(s_i - \mathtt{m}_{A \cup B}) + \sum_{i \in B} \exp(s_i - \mathtt{m}_{A \cup B}) \\
    &= \sum_{i \in A \cup B} \exp(s_i - \mathtt{m}_{A \cup B}) \\
\end{align*}
Expanding the computation of $\mathtt{w}_{A \cup B}$:
\begin{align*}
    \mathtt{w}_{A \cup B} &= \mathtt{w}_{A} \exp(\mathtt{m}_{A} - \mathtt{m}_{A \cup B}) + \mathtt{w}_{B} \exp(\mathtt{m}_{B} - \mathtt{m}_{A \cup B}) \\
    &= \left[\sum_{i \in A} \exp(s_i - \mathtt{m}_A) v_i \right]  \exp(\mathtt{m}_{A} - \mathtt{m}_{A \cup B}) + \left[\sum_{i \in B} \exp(s_i - \mathtt{m}_B)v_i \right] \exp(\mathtt{m}_{B} - \mathtt{m}_{A \cup B}) v_i \\
    &= \sum_{i \in A} \exp(s_i - \mathtt{m}_{A \cup B}) v_i + \sum_{i \in B} \exp(s_i - \mathtt{m}_{A \cup B})v_i \\
    &= \sum_{i \in A \cup B} \exp(s_i - \mathtt{m}_{A \cup B})v_i \\
\end{align*}
resulting in $\mathtt{m}_{A \cup B} = \mathrm{max}_{i \in A \cup B} s_i$, $\mathtt{u}_{A \cup B} = \sum_{i \in A \cup B} \exp(s_i - \mathtt{m}_{A \cup B})$, and $\mathtt{w}_{A \cup B} = \sum_{i \in A \cup B} \exp(s_i - \mathtt{m}_{A \cup B})v_i$ as needed.

\subsection{Associative Proof}
\label{appendix:parallel_scan:associative}

For the operator $\oplus$ to be valid for the parallel scan algorithm, the operator $\oplus$ must satisfy the associative property, i.e., $(\mathtt{a} \oplus \mathtt{b}) \oplus \mathtt{c} = \mathtt{a} \oplus (\mathtt{b} \oplus \mathtt{c})$.

Since the detailed operator is applied to $3$-tuples of the form $(\mathtt{m}_A, \mathtt{u}_A, \mathtt{w}_A)$ as follows:
$$
(\mathtt{m}_{A}, \mathtt{u}_{A}, \mathtt{w}_{A}) \oplus (\mathtt{m}_{B}, \mathtt{u}_{B}, \mathtt{w}_{B}) = 
(\mathtt{m}_{A \cup B}, \mathtt{u}_{A \cup B}, \mathtt{w}_{A \cup B})
$$
therefore showing that $\oplus$ is associative means that: 
\begin{align*}
\left[(\mathtt{m}_{A}, \mathtt{u}_{A}, \mathtt{w}_{A}) \oplus (\mathtt{m}_{B}, \mathtt{u}_{B}, \mathtt{w}_{B})\right] \oplus (\mathtt{m}_{C}, \mathtt{u}_{C}, \mathtt{w}_{C})
&=
(\mathtt{m}_{A}, \mathtt{u}_{A}, \mathtt{w}_{A}) \oplus \left[(\mathtt{m}_{B}, \mathtt{u}_{B}, \mathtt{w}_{B}) \oplus (\mathtt{m}_{C}, \mathtt{u}_{C}, \mathtt{w}_{C})\right] \\
(\mathtt{m}_{(A \cup B) \cup C}, \mathtt{u}_{(A \cup B) \cup C}, \mathtt{w}_{(A \cup B) \cup C})
&=
(\mathtt{m}_{A \cup (B \cup C)}, \mathtt{u}_{A \cup (B \cup C)}, \mathtt{w}_{A \cup (B \cup C)}) \\
\end{align*}
Since $\cup$ is an associative operator that acts on sets, i.e., $(A \cup B) \cup C = A \cup (B \cup C)$, therefore
$\mathtt{m}_{(A \cup B) \cup C} = \mathtt{m}_{A \cup (B \cup C)}$, $\mathtt{u}_{(A \cup B) \cup C} = \mathtt{u}_{A \cup (B \cup C)}$, and $\mathtt{w}_{(A \cup B) \cup C} = \mathtt{w}_{A \cup (B \cup C)}$, meaning that $\oplus$ is associative as required.

%% file: appendix/99_03_dataset_description.tex
\section{Breakdown of Individual Datasets}
\label{appendix:dataset}

\subsection{Reinforcement Learning}
\label{appendix:dataset:rl}

Our code for the reinforcement learning experiments is based on that of \citet{minimal_decision_transformer} (MIT License).
For these experiments, we consider locomotion MuJoCo robotics environments (Apache 2.0 License) popular in deep RL:
\begin{itemize}
    \item HalfCheetah is a two-legged $2$-d robot comprising of a head, horizontal torso, two thighs, two shins, and two feet. 
    \item Ant is a $3$-d four-legged robot comprising of a torso and four legs ($2$ parts per leg).
    \item Hopper is a single-legged $2$-d robot comprising a torso, thigh, shin, and foot.
    \item Walker(2D) is a two-legged $2$-d robot comprising of a vertical torso, two thighs, two shins, and two feet. 
\end{itemize}

The datasets used in our experiments are from D4RL~\citep{fu2020d4rl} (Apache 2.0 License): 
\begin{itemize}
    \item Medium is a dataset generated by a pre-trained Soft Actor-Critic policy~\citep{haarnoja2018soft} that was early stopped during training.
    \item Medium-Expert is a dataset comprising of equal parts: expert and suboptimal demonstrations.
    \item Medium-Replay is a dataset consisting of samples from the replay buffer during the training of the medium policy.
\end{itemize}

\subsection{Event Forecasting}
\label{appendix:dataset:event_forecasting}

Our code for the event forecasting experiments and datasets is based on that of \citet{bae2023meta} (CC BY-NC-SA 4.0 License).
For these experiments, we consider $8$ datasets used in prior works~\citep{bae2023meta,zhang2020self,zuo2020transformer}. 

$5$ out of the datasets include marks/labels:
\begin{itemize}
    \item MIMIC (MIMIC-II) is a dataset based on an electric medical record system, consisting of anonymous patients' clinical visits.
    \item Wiki is a dataset based on Wikipedia, consisting of actions performed on the most edited pages.
    \item Reddit is a dataset based on the social media platform, consisting of the actions of the most active users.
    \item Mooc is a dataset based on an online course, consisting of the sequence of actions performed by various users.
    \item StackOverflow is a dataset based on the question-answering website, consisting of user awards.
\end{itemize}

$3$ out of the $8$ datasets do not include marks/labels: 
\begin{itemize}
    \item Sin is a synthetic dataset generated from a sine function with a periodicity of $4\pi$ and a domain of $[0, 32\pi]$.
    \item Uber is a dataset based on Uber NYC pickup data from 2015.
    \item Taxi is a dataset based on NYC Taxi pickup data from 2013. 
\end{itemize}

\subsection{Time Series Forecasting}
\label{appendix:dataset:tsf}

Our code for the time series forecasting experiments is based on the Time series Library repository (MIT License) by \citet{wu2023timesnet}.
For these experiments, we consider $8$ popular datasets available in the Time Series Library repository:
\begin{itemize}
    \item Weather is a dataset containing meteorological indicators recorded every 10 minutes for 2020.
    \item Exchange is a dataset consisting of the daily exchange rates of 8 countries (Australia, British, Canada, Switzerland, China, Japan, New Zealand and Singapore) from 1990 to 2016.
    \item Traffic is a dataset consisting of the hourly occupancy rate of $96$3 car lanes of San Francisco bay area freeways.
    \item ECL is a dataset consisting of electricity consumption
of $321$ clients.
    \item ETT (h1, h2, m1, and m2) are datasets consisting of data from two stations' electricity transformers whereas "h" denotes the hourly measurement and "m" denotes the minute-by-minute measurement. 
\end{itemize}

\subsection{Time Series Classification}
\label{appendix:dataset:tsc}

Our code for the time series classification experiments is based on the Time series Library repository (MIT License) by \citet{wu2023timesnet}.
For these experiments, we consider $10$ UEA datasets~\citep{bagnall2018uea} available in the Time Series Library repository:
\begin{itemize}
    \item EthanolConcentration is a dataset of raw spectra recordings of whisky bottles and their alcohol concentration. 
    \item FaceDetection is a dataset of MEG (Magnetoencephalography) recordings and their class labels (Face or Scramble). 
    \item Handwriting is a dataset of measurements taken from a smart watch while writing. 
    \item Heartbeat is a dataset of heart sound recordings derived from the 2016 PhysioNet/Computing in Cardiology Challenge Challenge.
    \item JapaneseVowels is a dataset of recordings of Japanese-male vowel pronunciations. 
    \item PEMS-SF is a dataset describing the occupancy rate of car lanes in San Francisco Bay Area's freeways available on California's Department of Transportation PEMS (Performance Measurement System) website.
    \item SelfRegulationSCP1 is a dataset of cortical potential recordings from a healthy subject.
    \item SelfRegulationSCP2 is a dataset of cortical potential recordings from an artificially respirated ALS (Amyotrophic lateral sclerosis) patient.
    \item SpokenArabicDigits is a dataset of recordings of ten spoken Arabic digits from $44$ male and $44$ female Arabic native speakers.
    \item UWaveGestureLibrary is a dataset of accelerometer measurements of eight gestures.
\end{itemize}

%% file: appendix/99_04_experiments.tex
\section{Additional Experiments}
\label{appendix:experiments}

\subsection{Full Time Series Forecasting Results}

In time series forecasting, the objective of the model is to predict $T$ future values of the series. Following the standard of previous works~\citep{wu2023timesnet}, we evaluated on $t \in \{96, 192, 336, 720\}$. However, due to space limitations, only $T = 192$ was included in the main paper. Here, in Table \ref{appendix:table:tsf:full_results}, we include the full results. 
We see that for all values of $T$, Aarens performed comparably with Transformers across all datasets. However, unlike Transformers, Aarens can efficiently process new inputs, making it more advantageous in sequential settings such as this time series-related setting. 

\begin{table}[]
\centering

\bgroup
\def\arraystretch{1.25}
\begin{tabular}{c|c|cc|cc}
\hline
\multicolumn{6}{c}{\large \textbf{Time Series Forecasting}}                                                                                                 \\ \hline
\multicolumn{2}{c}{Metric}      & \multicolumn{2}{|c}{MSE ($\downarrow$)}              & \multicolumn{2}{|c}{MAE ($\downarrow$)}              \\ \hline
\multicolumn{2}{c}{Models}      & \multicolumn{1}{|c}{Aaren}                    & Transformer              & Aaren                    & Transformer              \\ \hline
\multirow{4}{*}{ETTh1}    & $96$  & $\mathbf{0.53 \pm 0.04}$ & $\mathbf{0.54 \pm 0.01}$ & $\mathbf{0.52 \pm 0.02}$ & $\mathbf{0.50 \pm 0.01}$ \\
                          & $192$ & $\mathbf{0.59 \pm 0.03}$ & $\mathbf{0.64 \pm 0.05}$ & $\mathbf{0.55 \pm 0.01}$ & $\mathbf{0.57 \pm 0.02}$ \\
                          & $336$ & $\mathbf{0.65 \pm 0.03}$ & $\mathbf{0.65 \pm 0.02}$ & $\mathbf{0.55 \pm 0.01}$ & $\mathbf{0.55 \pm 0.01}$ \\
                          & $720$ & $\mathbf{0.67 \pm 0.05}$ & $\mathbf{0.70 \pm 0.05}$ & $\mathbf{0.62 \pm 0.02}$ & $\mathbf{0.58 \pm 0.02}$ \\ \hline
\multirow{4}{*}{ETTh2}    & $96$  & $\mathbf{0.38 \pm 0.02}$ & $\mathbf{0.41 \pm 0.04}$ & $\mathbf{0.44 \pm 0.02}$ & $\mathbf{0.40 \pm 0.02}$ \\
                          & $192$ & $\mathbf{0.49 \pm 0.03}$ & $\mathbf{0.50 \pm 0.03}$ & $\mathbf{0.48 \pm 0.02}$ & $\mathbf{0.46 \pm 0.01}$ \\
                          & $336$ & $\mathbf{0.57 \pm 0.05}$ & $\mathbf{0.59 \pm 0.03}$ & $\mathbf{0.47 \pm 0.02}$ & $\mathbf{0.50 \pm 0.01}$ \\
                          & $720$ & $\mathbf{0.55 \pm 0.03}$ & $\mathbf{0.60 \pm 0.03}$ & $\mathbf{0.52 \pm 0.01}$ & $\mathbf{0.52 \pm 0.01}$ \\ \hline
\multirow{4}{*}{ETTm1}    & $96$  & $0.48 \pm 0.02$          & $\mathbf{0.44 \pm 0.01}$ & $0.44 \pm 0.01$          & $\mathbf{0.41 \pm 0.01}$ \\
                          & $192$ & $\mathbf{0.51 \pm 0.03}$ & $\mathbf{0.52 \pm 0.05}$ & $\mathbf{0.47 \pm 0.01}$ & $\mathbf{0.47 \pm 0.01}$ \\
                          & $336$ & $\mathbf{0.54 \pm 0.02}$ & $\mathbf{0.57 \pm 0.03}$ & $\mathbf{0.49 \pm 0.01}$ & $\mathbf{0.51 \pm 0.01}$ \\
                          & $720$ & $\mathbf{0.60 \pm 0.03}$ & $\mathbf{0.66 \pm 0.06}$ & $\mathbf{0.52 \pm 0.01}$ & $\mathbf{0.56 \pm 0.02}$ \\ \hline
\multirow{4}{*}{ETTm2}    & $96$  & $\mathbf{0.24 \pm 0.03}$ & $\mathbf{0.25 \pm 0.01}$ & $\mathbf{0.30 \pm 0.02}$ & $\mathbf{0.30 \pm 0.01}$ \\
                          & $192$ & $\mathbf{0.34 \pm 0.04}$ & $\mathbf{0.38 \pm 0.02}$ & $\mathbf{0.39 \pm 0.02}$ & $\mathbf{0.37 \pm 0.01}$ \\
                          & $336$ & $\mathbf{0.41 \pm 0.03}$ & $\mathbf{0.49 \pm 0.05}$ & $\mathbf{0.42 \pm 0.01}$ & $\mathbf{0.43 \pm 0.02}$ \\
                          & $720$ & $\mathbf{0.51 \pm 0.03}$ & $\mathbf{0.56 \pm 0.02}$ & $\mathbf{0.49 \pm 0.02}$ & $\mathbf{0.47 \pm 0.01}$ \\ \hline
\multirow{4}{*}{Weather}  & $96$  & $\mathbf{0.18 \pm 0.00}$ & $\mathbf{0.18 \pm 0.00}$ & $\mathbf{0.23 \pm 0.00}$ & $\mathbf{0.23 \pm 0.00}$ \\
                          & $192$ & $\mathbf{0.25 \pm 0.01}$ & $\mathbf{0.24 \pm 0.01}$ & $\mathbf{0.28 \pm 0.00}$ & $\mathbf{0.28 \pm 0.00}$ \\
                          & $336$ & $\mathbf{0.31 \pm 0.00}$ & $\mathbf{0.31 \pm 0.02}$ & $\mathbf{0.32 \pm 0.00}$ & $0.34 \pm 0.01$          \\
                          & $720$ & $\mathbf{0.40 \pm 0.00}$ & $\mathbf{0.38 \pm 0.02}$ & $\mathbf{0.39 \pm 0.00}$ & $\mathbf{0.39 \pm 0.01}$ \\ \hline
\multirow{4}{*}{Exchange} & $96$  & $\mathbf{0.14 \pm 0.01}$ & $\mathbf{0.14 \pm 0.01}$ & $\mathbf{0.27 \pm 0.01}$ & $\mathbf{0.25 \pm 0.01}$ \\
                          & $192$ & $\mathbf{0.25 \pm 0.03}$ & $\mathbf{0.24 \pm 0.02}$ & $\mathbf{0.33 \pm 0.02}$ & $\mathbf{0.34 \pm 0.01}$ \\
                          & $336$ & $\mathbf{0.42 \pm 0.04}$ & $\mathbf{0.41 \pm 0.02}$ & $\mathbf{0.44 \pm 0.02}$ & $\mathbf{0.45 \pm 0.01}$ \\
                          & $720$ & $\mathbf{1.20 \pm 0.07}$ & $\mathbf{1.44 \pm 0.19}$ & $\mathbf{0.79 \pm 0.02}$ & $\mathbf{0.81 \pm 0.04}$ \\ \hline
\multirow{4}{*}{Traffic}  & $96$  & $\mathbf{0.63 \pm 0.01}$ & $\mathbf{0.61 \pm 0.01}$ & $0.35 \pm 0.00$ & $\mathbf{0.34 \pm 0.00}$ \\
                          & $192$ & $\mathbf{0.64 \pm 0.01}$ & $\mathbf{0.63 \pm 0.01}$ & $0.35 \pm 0.00$ & $\mathbf{0.34 \pm 0.00}$ \\
                          & $336$ & $\mathbf{0.65 \pm 0.01}$ & $\mathbf{0.64 \pm 0.00}$ & $0.35 \pm 0.00$ & $\mathbf{0.34 \pm 0.00}$ \\
                          & $720$ & $\mathbf{0.68 \pm 0.01}$ & $\mathbf{0.67 \pm 0.00}$ & $\mathbf{0.36 \pm 0.01}$ & $\mathbf{0.36 \pm 0.00}$ \\ \hline
\multirow{4}{*}{ECL}      & $96$  & $\mathbf{0.36 \pm 0.02}$ & $\mathbf{0.35 \pm 0.02}$ & $0.46 \pm 0.01$ & $\mathbf{0.43 \pm 0.01}$ \\
                          & $192$ & $\mathbf{0.37 \pm 0.02}$ & $\mathbf{0.39 \pm 0.03}$ & $\mathbf{0.45 \pm 0.01}$ & $\mathbf{0.48 \pm 0.02}$ \\
                          & $336$ & $\mathbf{0.47 \pm 0.05}$ & $\mathbf{0.48 \pm 0.06}$ & $\mathbf{0.52 \pm 0.03}$ & $\mathbf{0.55 \pm 0.03}$ \\
                          & $720$ & $\mathbf{0.57 \pm 0.05}$ & $\mathbf{0.62 \pm 0.06}$ & $\mathbf{0.56 \pm 0.02}$ & $\mathbf{0.55 \pm 0.03}$ \\ \hline
\end{tabular}
\egroup
\caption{Full Time Series Forecasting Results.}
\label{appendix:table:tsf:full_results}
\end{table}

%% file: appendix/99_05_implementation-details.tex
\section{Hyperparameters + Implementation Details}
\label{appendix:implementation_details}

As Transformers and Aarens share the same interface and are both attention-based methods, they share the same set of hyperparameters. 
For fairness, the same hyperparameters are used for both Transformers and Aarens. 
The specific hyperparameters for their respective settings are as follows:

\textbf{Reinforcement Learning.} For these experiments, we found that hyperparameters used by \citet{zheng2022online} performed better than the default Decision Transformers hyperparameters. As such, \citet{zheng2022online}'s hyperparameters were used for these experiments, i.e., an embedding dimension of $512$, $4$ attention heads, and $4$ Transformers/Aarens blocks. 

\textbf{Event Forecasting.} For these experiments, the default hyperparameters (except for the learning rate) in \citet{bae2023meta}'s repository were used. 
We found that the default learning rate value of $0.0001$ caused early convergences to poor local optima.
As such, the learning rate was set to $0.0005$ instead. 

\textbf{Time Series Forecasting.} For these experiments, the default Transformers hyperparameters in the Time Series Library repository~\citep{wu2023timesnet} were used.

\textbf{Time Series Classification.} For these experiments, the default Transformers hyperparameters in the Time Series Library repository~\citep{wu2023timesnet} were used.

%% file: appendix/99_06_compute-resources.tex
\section{Compute}
\label{appendix:compute}

Our experiments were run using a mix of Nvidia GTX 1080 Ti (12 GB) and Nvidia Tesla P100 (16 GB) GPUs. The analyses were performed on Nvidia GTX 1080 Ti (12 GB). 

The reinforcement learning experiments required approximately the same amount of time: $\sim 2-4$ hours each.

The event forecasting experiments varied in time depending on the dataset:
\begin{itemize}
    \item MIMIC took $\sim 0.5$ hours
    \item Wiki took $\sim 0.75$ hours
    \item Reddit took $\sim 3.5$ hours
    \item Mooc took $\sim 8$ hours
    \item StackOverflow took $\sim 3.5$ hours
    \item Sin took $\sim 1.5$ hours
    \item Uber took $\sim 3$ hours
    \item Taxi took $\sim 1.5$ hours
\end{itemize}

The time series forecasting experiments were run as a single script for $T \in \{96, 192, 336, 720\}$. The experiments varied in time depending on the dataset:

\begin{itemize}
    \item Weather took $\sim 6$ hours
    \item Exchange took $\sim 0.5$ hours
    \item Traffic took $\sim 1$ hours
    \item ECL took $\sim 4$ hours
    \item ETTh1 took $\sim 0.75$ hours
    \item ETTm1 took $\sim 11$ hours
    \item ETTh2 took $\sim 0.75$ hours
    \item ETTm2 took $\sim 11$ hours
\end{itemize}

The time series classification experiments were run as a single script. Running the experiments on all datasets took in total $\sim 1$ hour.

%% file: appendix/99_07_limitations_broader-impact.tex
\section{Limitations and Broader Impact}
\label{appendix:limitations_broader_impact}

\textbf{Limitations.} Unlike Transformers whose attention queries are input-dependent (i.e., $x_i$), Aarens' attention queries are input-independent (i.e., $q$ is a learned constant). 
In our experiments (reinforcement learning, event forecasting, time series forecasting, and time series classification), we found that Aarens performed comparably to Transformers regardless. 
However, this difference could be a limitation in settings that require large highly expressive sequence models, e.g., large language models. 
This, however, is outside the scope of our work and our available resources.

\textbf{Broader Impact.} 
In this work, we propose Aaren, a module that achieves comparable performance to Transformers while being more time and memory-efficient. 
Since Aarens can update themselves efficiently (as an RNN) unlike Transformers, therefore Aarens are more suitable for sequence modelling.
Since (1) there exists a wide range of sequence modelling problem settings and (2) the number of low-resource domains (e.g., battery powered or IoT devices) is rapidly increasing, therefore Aaren has a broad potential impact. 
The general applicability of Aaren means that its societal impact is dependent on the downstream application.